\documentclass{article}
\usepackage[toc,page]{appendix}
\usepackage{spconf,amsmath,graphicx}
\usepackage{booktabs}
\usepackage{cite}
\usepackage{graphicx}
\usepackage{hyperref}
\usepackage{multirow}
\usepackage{textcomp}
\usepackage{xcolor}
\usepackage{algorithmic}
\usepackage{adjustbox}
\usepackage{enumitem}
\usepackage[binary-units=true,version-1-compatibility]{siunitx}
\usepackage[colorinlistoftodos]{todonotes}

\DeclareMathOperator{\stddev}{stddev}
\DeclareMathOperator{\mean}{mean}
\DeclareMathOperator{\range}{range}
\title{Replacing Human Audio with Synthetic Audio\\for On-device Unspoken Punctuation Prediction}
\name{Daria Soboleva, Ondrej Skopek, M\'{a}rius \v{S}ajgal\'{i}k, Victor C\u{a}rbune, Felix Weissenberger}
\secondlinename{Julia Proskurnia, Bogdan Prisacari, Daniel Valcarce, Justin Lu, Rohit Prabhavalkar, Balint Miklos}
\address{
Google, Z\"{u}rich, Switzerland\\
\texttt{$\{$dariasobol, oskopek, sajgalik, vcarbune$\}$@google.com}
}

\begin{document}
%
\maketitle
{
\let\thefootnote\relax\footnotetext{%
\hspace{-6mm} Accepted to \href{https://2021.ieeeicassp.org/}{IEEE ICASSP 2021}.\\
\\
\copyright~2021 IEEE.  Personal use of this material is permitted.  Permission from IEEE must be obtained for all other uses, in any current or future media, including reprinting/republishing this material for advertising or promotional purposes, creating new collective works, for resale or redistribution to servers or lists, or reuse of any copyrighted component of this work in other works.}
}
\begin{abstract}
We present a novel multi-modal unspoken punctuation prediction system for the English language which combines acoustic and text features. We demonstrate for the first time, that by relying exclusively on synthetic data generated using a prosody-aware text-to-speech system, we can outperform a model trained with expensive human audio recordings on the unspoken punctuation prediction problem.
Our model architecture is well suited for on-device use. This is achieved by leveraging hash-based embeddings of automatic speech recognition text output in conjunction with acoustic features as input to a quasi-recurrent neural network, keeping the model size small and latency low.

\end{abstract}
\begin{keywords}
unspoken punctuation, acoustic features, speech processing
\end{keywords}
\section{Introduction}
\label{sec:intro}

Recent advances in on-device automatic speech recognition (ASR) technology~\cite{he2018streaming} enable a range of applications where speech recognition is central to the user experience, such as voice dictation or live captioning~\cite{livecaptionblog}.

A severe limitation of many ASR systems is the lack of high quality punctuation. ASR systems usually do not predict any punctuation symbols which have not been explicitly spoken out, which makes ASR-transcribed text hard to read and understand for users.

In this paper, we introduce a novel unspoken punctuation prediction system for refining ASR output. 
For example, given the input \emph{``hey Anna how are you''} together with roughly corresponding audio segments, our system converts it to \emph{``Hey, Anna! How are you?''}. This is in contrast to spoken punctuation prediction, where spoken inputs such as \emph{``hey comma Anna exclamation mark''} are converted into punctuation symbols \emph{``Hey, Anna!''}.

Several methods for unspoken punctuation prediction have been proposed previously. These can be categorized based on input features: either relying on acoustic (prosodic)
features~\cite{batista2012bilingual}, text features~\cite{cyberpunc, tilk2016bidirectional} or on a multi-modal approach combining both~\cite{klejch2017sequence}.

Approaches relying only on text features suffer from lack of quality, especially for utterances with ambiguous punctuation which heavily depend on prosodic cues. Conversely, relying on audio requires expensively annotated human audio recordings.
Maintaining data quality across numerous speakers when collecting large amounts of human audio is often very costly and presents a roadblock for training larger models --- even more so as punctuation marks suffer from semantic drift over time~\cite{EvolvinPunct1, EvolvinPunct2}.

In this paper, we combine text and acoustic features. To mitigate the scarcity of human recordings, we propose using synthetic audio, which allows us to obtain larger datasets and potentially train models for domains where no human audio is currently available, but a text-only dataset exists.
Given the recent success of using text-to-speech (TTS) models to train ASR systems~\cite{TTS-ASR-helper1, TTS-ASR-helper3, TTS-ASR-helper4, TTS-ASR-helper5}, we explore synthetic audio generation approaches for punctuation prediction using prosody-aware Tacotron TTS models~\cite{skerryryan2018endtoend}.

Firstly, we show that the quality of a punctuation model trained on \num{100}\% human audio can be matched using only \num{20}\% of the expensively collected human audio (LibriTTS). In addition, we show that replacing the dataset entirely with synthesized audio using multiple TTS speakers outperforms models trained only on human audio.

Our contributions are as follows:
\begin{itemize}
    \item Introduce a \textbf{novel multi-modal acoustic neural network architecture with low memory and latency} footprints suitable for on-device use.
    \item Achieve \textbf{superior performance of acoustic and text based models compared to the text-only baseline} with only a small increase in model parameter size.
    \item Using only \textbf{TTS-generated audio samples}, we outperform models trained on expensively collected human recordings and annotations.
\end{itemize}
To the best of our knowledge, this is the first approach that successfully replaces human audio with synthetic audio for punctuation prediction.

\section{Model Architecture}
\label{sec:model_architecture}

\begin{figure}[t]
\centering
\hspace*{-0.25cm}\includegraphics[scale=0.45]{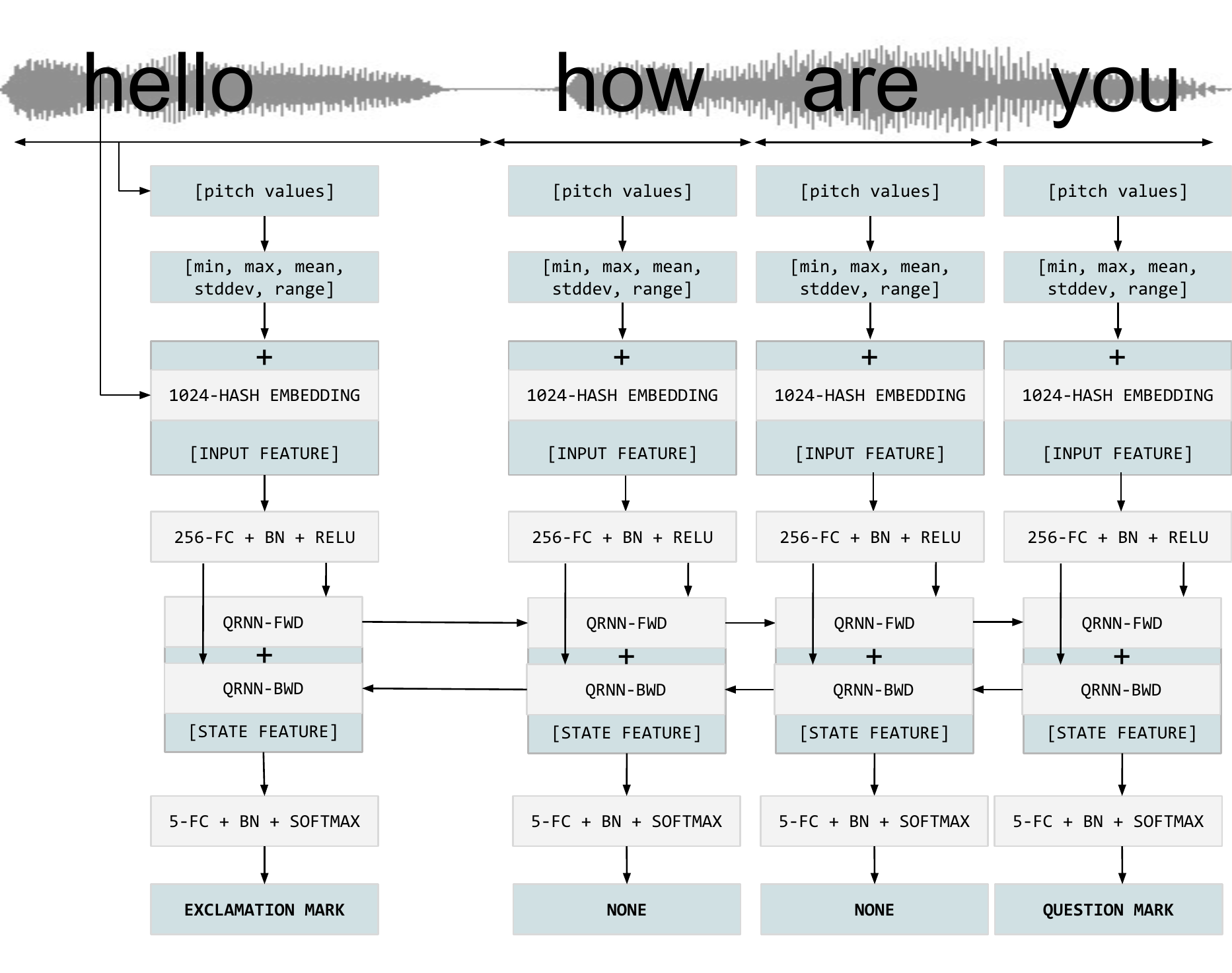}
\caption{Our multi-modal architecture on an example utterance. We concatenate text and acoustic features (``+'') and pass them through a bidirectional QRNN layer, followed by a fully-connected batch-normalized layer with softmax activation, classifying which punctuation symbol to append.}
\label{fig:modelstack}
\end{figure}

The model uses both extracted pitch values from audio and the output of the ASR system having processed the input audio. The ASR output consists of the utterance transcription, together with approximate times in the utterance when a new token is emitted.

The boundaries only roughly represent transitions from one token in the hypothesis to the next one and cannot be used to directly estimate pause lengths, because common ASR systems (e.g.~sequence-to-sequence~\cite{he2018streaming} or connectionist temporal
classification~\cite{graves2014towards}) do not provide exact token alignments, or they may not be necessarily accurate at inference time. In our work, we use the on-device ASR system described in~\cite{he2018streaming}.

\subsection{Text Features using Hash-based Embeddings}
Representations for individual tokens are determined using a ``hash-based embedding'',
computed on the fly through a locality-sensitive string hashing operation that relies on the byte-level
representation of the token, transforming it into a fixed dimensional representation of a requested size \cite{kaliamoorthi-etal-2019-prado}.

One of the main motivations for using hash-based embeddings is reducing the model size. 
Not having to store a learned matrix of dense embeddings for a vocabulary is advantageous for on-device use cases. For a relatively small vocabulary size of \num[sepfour]{20000}~tokens, even assuming a quantized \num{8}-bit integer representation, the resulting matrix of commonly used \num{128}-dimensional embeddings would increase the model size by roughly \SI{2.4}{\mebi\byte}.
Our models use roughly \SI{1}{\mebi\byte}, meaning this would triple their size.

\subsection{Pitch Estimation for Acoustic Features}

Representations for audio segments corresponding to tokens are computed using pitch estimation~\cite{batista2012bilingual}, with the YIN algorithm~\cite{de2002yin}, together with a voiced or unvoiced estimation as described in~\cite{talkin1995robust}. This method provides pitch estimates (frequencies in \si{\Hz}) for every \SI{5}{\ms} of input audio for voiced segments.
Assuming a sample rate of \SI{16}{\kHz}, we estimate \num{1}~pitch value per \num{80}~audio samples. The estimation method runs on the entire audio corresponding to the input utterance and the estimated pitch values are aligned using time-boundaries to the corresponding input tokens. Although ASR timings do not capture pauses between the tokens explicitly, the estimation method produces zero values for unvoiced segments.

The token-aligned pitch values are then used to compute acoustic features per token. We used \num{5} scalar statistics as acoustic features: $\mean$, $\stddev$, $\max$, $\min$ and $\range$ (absolute difference between $\max$ and $\min$). The vector of computed acoustic scalar statistics are then concatenated with text embeddings and further used as input features~(Fig.~\ref{fig:modelstack}).

\subsection{Quasi-Recurrent Neural Network Layers}
Before passing input features into a time-dependent layer, we project them down
to a \num{256}-dimensional vector using a fully-connected batch-normalized layer with ReLU activation. 

The sequence of projected feature vectors is then passed through a bidirectional quasi-recurrent neural network layer~\cite{bradbury2016quasirecurrent}. We apply dropout on the hidden units of this layer, referred to as a zoneout~\cite{krueger2016zoneout}.
The convolution-based QRNN architecture runs independently for each time-step and uses a time-dependent pooling layer across channels. In contrast to regular RNN layers, all computations except for the lightweight pooling, do not depend on previous time-steps, enabling faster inference.

Finally, the per-token outputs are concatenated and passed into a softmax classification layer. At training time, we compute a cross-entropy loss with $\ell_2$ regularization over model weights using the output probabilities for \num{5} classes: \textsc{Period}, \textsc{Question Mark}, \textsc{Exclamation Mark}, \textsc{Comma} and \textsc{None}. Each class, except for \textsc{None}, is mapped to the corresponding punctuation symbol appended to the input token.

This approach allows us to infer punctuation for sequences up to \num{100} tokens long in less than \SI{5}{\ms}.

\section{Methods \& Experiments}
\label{sec:experiments}

\subsection{Data}
\label{datasets}
\begin{table*}[t]
\centering
\caption{Number of samples in the training, validation, and test splits for the LibriTTS dataset.
For the test split, frequency counts of labels are also provided.
}
\label{datatable}

\tabcolsep=0.11cm
\begin{tabular}{rrrrrrrrrrrr}
\toprule \multicolumn{1}{c}{\textbf{Train}}&\phantom{.}&\multicolumn{1}{c}{\textbf{Validation}}&\phantom{.}&\multicolumn{8}{c}{\textbf{Test}} \\
\cmidrule{1-1}
\cmidrule{3-3}
\cmidrule{5-12}
 samples &&   samples &&   samples & tokens &
 punct. &
 \textsc{EoS} &
 \textsc{Period}  &
 \textsc{Question Mark} & 
 \textsc{Exclamation Mark} &
 \textsc{Comma} \\ \midrule
 \num[sepfour]{42773} &&    \num[sepfour]{2253} && \num[sepfour]{1675}  &    \num[sepfour]{24973} & \num[sepfour]{3206}  &  \num[sepfour]{1683} &       \num[sepfour]{1520}  &  \num[sepfour]{97} &  \num[sepfour]{66}   &     \num[sepfour]{1523} \\
\bottomrule
\end{tabular} 
\end{table*}

\begin{table*}[t]
\centering
\caption{Models are evaluated on LibriTTS test set with human audio. We report punctuation token accuracy for the dataset, F1-scores for individual punctuation symbols and for end-of-sentence (\textsc{EoS}).
Human / TTS \ indicates the percentage of text samples augmented with each audio type. Metrics are given in percentages and averaged over \num{3} runs. Best results are in bold.}
\begin{tabular}{cccccccc}
\toprule
\multirow{2.5}{*}{\textbf{Human / TTS}} &\multirow{2.5}{1.85cm}{\textbf{Punctuation \hphantom{n}Accuracy}} & & \multicolumn{4}{c}{\textbf{F1}} \\
\cmidrule{3-7} 
    & & \textsc{EoS} & \textsc{Period} & \textsc{Question mark} & \textsc{Exclamation mark} & \textsc{Comma} \\\midrule
\hphantom{10}\num{0} / \hphantom{00}\num{0} & \num{76.35} &  \num{96.90} & \num{94.42} & \num{72.62} & \pmb{\num{28.19}}  & \num{51.40}\\\midrule

\num{100} / \hphantom{00}\num{0} & \num{81.56} &   \num{96.87} & \num{94.09} & \num{72.87} & \num{22.20}  & \pmb{\num{59.76}}\\
\hphantom{1}\num{50} / \hphantom{0}\num{50} & \num{81.67} &  \num{96.94} & \num{94.38} & \num{74.75} & \num{23.52}  & \pmb{\num{59.35}}\\
\hphantom{1}\num{20} / \hphantom{0}\num{80} & \num{81.38} &  \num{96.76} & \num{94.21} & \pmb{\num{77.49}} & \num{22.19}  & \pmb{\num{58.02}} \\
\hphantom{10}\num{0} / \num{100} & \num{81.37} &  \num{96.78} & \num{93.90} & \num{73.40} & \num{18.39}  & \num{57.25}\\
\hphantom{10}\num{0} / \num{200}   & \pmb{\num{86.15}} &  \num{96.62} & \num{94.17} & \num{76.04} & \pmb{\num{29.64}}  & \num{56.22}\\

\bottomrule
\end{tabular}
\label{HumanTTSAudioMix}
\end{table*}

We use the publicly available LibriTTS corpus\footnote{\url{https://www.openslr.org/60}} introduced in~\cite{zen2019libritts}, which consists of \num{585}~hours of English human speech, downsampled to \SI{16}{\kHz}.

The dataset is derived from LibriSpeech~\cite{librispeech2005}. It preserves original text including punctuation, splits speech at sentence boundaries, and utterances with significant background noise are removed.
In order to replace human audio with synthetic audio, we augmented the training samples with the corresponding synthetically generated audio. The audio is synthesized using a Tacotron model, trained on a multi-speaker dataset as described in~\cite{skerryryan2018endtoend}.

\subsection{Data Augmentation with TTS}
One of the natural advantages of synthetic TTS audio is that we can generate a large amount of it automatically given a text corpus and a TTS model.
Hence, we propose a data augmentation technique: given a text sample, generate multiple synthetic audio samples with voices of different speakers. By doing so, we introduce additional audio samples improving the training data representation. With this approach, we can augment training data $N$-fold, $N$ being the maximum number of speakers available for a given TTS model.

\subsection{Preprocessing Details}

\subsubsection{Audio --- Text-to-Speech Generation}

Using Tacotron, we generate audio utterances corresponding to the preprocessed text samples and replace the original human audio with the generated version. We experimented with \num{52}~TTS English speakers in total. For model training and validation, we randomly split speakers into \num{90}\% training and \num{10}\% validation samples. When utilizing human LibriTTS utterances, we keep the original training, validation, and test split~\cite{zen2019libritts} (Table~\ref{datasets}). For testing, we  use the original human audio.

\subsubsection{Text --- Punctuation-aware Preprocessing}

The raw text is first tokenized, separating whitespace, punctuation and word tokens based on Unicode word
boundaries\footnote{\url{https://unicode.org/reports/tr29/\#Word\_Boundaries}}. Subsequently, the tokens are organized into sentences based on end-of-sentence punctuation (\textsc{Period}, \textsc{Question Mark}, \textsc{Exclamation Mark}) to form samples. A sample can contain multiple sentences as long as the total number of tokens in it is at least \num{3} and at most \num{100}, and contains at least \num{1}~punctuation mark.

\subsubsection{Audio --- Speaker and ASR Misrecognition Filtering}

Speech recognition systems are not perfect, and misrecognitions need to be dealt with. For training, we drop samples for which the number of ASR-recognized tokens is not equal to the number of tokens in the ground truth transcription, as for these it would be non-trivial to reconstruct the correct punctuation.
During this stage, we filter out roughly \SI{30}{\percent} samples.

\subsection{Training Details}
\subsubsection{Weighted Cross-Entropy Loss}
The distribution of punctuation in text is highly nonuniform (Table~\ref{datatable}). To mitigate this, we introduce a class-weighted cross-entropy loss with weights equal to the inverse of punctuation mark occurrence frequency for each class, calculated on the training split~\cite{weighted-ce}. This modification of the loss function ensures that tasks with extensive data available (such as \textsc{Period} or \textsc{Comma}) do not overly dominate the training, resulting in poor performance on classes with small amount of data (such as \textsc{Question Mark} or \textsc{Exclamation Mark}).
In-line with PRADO~\cite{kaliamoorthi-etal-2019-prado}, we add a small $L_2$~weight regularizer term to our loss with a multiplicative scale of 
\num{e-5}.

\subsubsection{Model Hyperparameters}
All models use a hash-based embedding dimension of \num[sepfour]{1024}, QRNN convolution kernel width of \num{7}, hidden state size of \num{80} and zone-out probability of \num{0.1}.
We train using Adam~\cite{adam} with an initial learning rate of~\num{5e-4} which is exponentially decayed by~\num{0.5} every \num[sepfour]{5000}~steps.
All models are trained for a total of \num[sepfour]{30000}~steps with batch size~\num{512}.
Text-only models have \num[sepfour]{838127} trainable parameters,
which increases for text-acoustic models by only \SI{0.15}{\percent} to \num[sepfour]{839407}.

\section{Results}
\label{results}

\subsection{Comparing Acoustic \& Text-only Models}
To motivate the use of acoustic features, we first compare the text-only model (\SI{0}{\percent} Human, \SI{0}{\percent} TTS) with an acoustic model that utilizes only human audio (\SI{100}{\percent} Human). The results in Table~\ref{HumanTTSAudioMix} show that the acoustic model outperforms the text-only model in punctuation accuracy and across most individual punctuation mark F1-scores. \textsc{Comma} prediction quality is significantly better, which demonstrates the capability of the model to implicitly infer pauses, despite explicit segmentation from ASR not being present. There is some quality degradation for \textsc{Exclamation Mark}, which the model mostly confused with \textsc{Period}. We address this issue in followup experiments.

\subsection{Human \& TTS Audio Combinations}
Given the abundance of text data, we can train large models for punctuation prediction.
However, with the lack of human audio, we cannot utilize acoustic features.
To mitigate this issue, we propose partially replacing human audio with synthetic TTS audio to find out how much human audio is necessary to maintain the same model performance.
To obtain a mix with the desired percentage of each audio type, during preprocessing, we perform sampling from the appropriate Bernoulli distribution for every text instance to decide which audio type to use for that sample.

In Table~\ref{HumanTTSAudioMix}, we see that replacing \SI{100}{\percent} of human audio with \SI{100}{\percent} TTS audio suffers from lower quality \textsc{Comma} predictions. 
By keeping just \SI{20}{\percent} human audio we observe a significant improvement for \textsc{Comma}, and the 20/80 and 50/50 experiments match performance of 100/0 (human audio). F1-score is better for \textsc{Question Mark}, likely due to increased variance from using additional speakers.

Hence, one could have annotated only \SI{20}{\percent} of the LibriTTS corpus with human audio and utilized TTS audio for the remainder of the data, without sacrificing punctuation model quality.

Models with \SI{50}{\percent} human audio show even better results for almost all punctuation marks.
Note that mixing of different audio types produces consistently better punctuation models, likely due to great data diversity in audio space --- e.g.~classes such as \textsc{Exclamation mark} or \textsc{Question mark} are often mistaken for \textsc{Period} because of pronunciation.
Relying on different speakers increases the chance of obtaining more general training data, improving model quality. In our experiments, mixes with \SI{40}{\percent}, \SI{60}{\percent}, and \SI{80}{\percent} human audio do not yield further improvements.

\subsection{Data Augmentation with Text-to-Speech Audio}

To reduce the reliance on human recorded and annotated audio, we synthetically generate all audio in the dataset. For each sample
we select at random $2$ out of the $52$ TTS-speakers to generate two audio variants of the same text sample. This enables us to
obtain a \SI{200}{\percent} larger dataset compared to the original one, with greater speaker variance.

Table~\ref{HumanTTSAudioMix} shows that models trained on this synthetic datasets outperform models trained on human audio
in punctuation accuracy and F1-score for \textsc{Exclamation Mark}. When using human audio, this class is commonly mistaken
with \textsc{Period} and the quality strictly depends on how human speakers pronounce sentences with it.

By leveraging a multi-speaker prosody-aware TTS model, the synthetic dataset proves to better represent acoustic changes relevant for punctuation prediction. The model training incorporates more variability in the feature space when using the synthetic dataset, compared to the original LibriTTS dataset, resulting in better performance. In our experiments, sampling more than \num{2} speakers per text sample does not yield additional improvements.

\section{Discussion and Conclusion}
\label{sec:conclusion}

In this paper, we propose a multi-modal punctuation prediction system that utilizes both text and acoustic features, while maintaining a practical setup that can run on-device, with low memory and latency footprints.
According to our experiments on LibriTTS, only \SI{20}{\percent} text data needs to be paired with human recordings and the rest can be synthesized by a TTS model without any quality loss.
Additionally, models trained using our TTS data augmentation with multiple speakers outperform models trained on human audio.
This allows us to alleviate the inherent data scarcity limitation of acoustic-based systems
and enable training punctuation models for any language for which a big-enough text corpus and a TTS model are available. In future work, we aim to train a TTS model on in-domain data (audiobooks), investigate differences due to high variance of human speech, and train on larger datasets.


\bibliographystyle{IEEEbib}
\bibliography{refs}

\vfill\pagebreak

\begin{appendices}
\section{Enhanced Acoustic Features}
\label{Appendix_A}

\begin{table*}[t]
\centering
\caption{Extraction of different acoustic features. Training on \SI{200}{\percent} TTS generated audio. Evaluation on LibriTTS test set with human audio. We report punctuation token accuracy for the dataset, F1-scores for each individual punctuation symbols and for end-of-sentence (\textsc{EoS}), where \textsc{Period}, \textsc{Exclamation Mark}, and \textsc{Question Mark} are considered interchangeable. Metrics are given in percentages and averaged over \num{3} runs. Best results are in bold.}

\begin{tabular}{cccccccc}
\toprule
\multirow{2.5}{*}{\textbf{Acoustic Feature}} &\multirow{2.5}{1.85cm}{\textbf{Punctuation \hphantom{n}Accuracy}} & & \multicolumn{4}{c}{\textbf{F1}} \\
\cmidrule{3-7} 
    & & \textsc{EoS} & \textsc{Period} & \textsc{Question mark} & \textsc{Exclamation mark} & \textsc{Comma} \\\midrule
Pitch-YIN & \num{86.15} &  \pmb{\num{96.62}} & \pmb{\num{94.17}} & \pmb{\num{76.04}} & \pmb{\num{29.64}}  & \pmb{\num{56.22}}\\\midrule

Pitch-SPICE & \num{81.54} &  \pmb{\num{96.70}} & \pmb{\num{94.01}} & \num{72.60} & \pmb{\num{29.35}}  & \num{55.30}\\

Log-mel & \pmb{\num{91.02}} &  \num{95.34} & \num{92.68} & \num{63.16} & \num{24.74}  & \num{41.04}\\

\bottomrule
\end{tabular}
\label{EnhancedAcousticFeatures}
\end{table*}

To improve the quality of our best models trained on \SI{200}{\percent} TTS synthetic audio (Table~\ref{HumanTTSAudioMix}), we experimented with different types of acoustic features.

First, we assess a different pitch estimation technique which relies on a pre-trained version of the SPICE model\footnote{\url{https://tfhub.dev/google/spice/1}} described in~\cite{spice}. The model outputs relative normalized pitch changes together with an uncertainty value for every \SI{32}{\ms} of audio (every \num{512}-samples for \SI{16}{\kHz} audio). We only rely on the normalized pitch change estimates for experiments outlined in this paper. The pre-trained SPICE model size is roughly \SI{1.8}{\mebi\byte}, which increases the size of the on-device stack compared to the YIN-based estimation~\cite{de2002yin} used previously.

Another type of acoustic features we extract from audio are \num{128}-dimensional log-mel filterbank energies. Log-mel features are a commonly used in speech processing systems~\cite{he2018streaming,skerryryan2018endtoend}. Moreover, they capture a richer voice signal representation compared to pitch values. Thus, we propose extracting more high-level features by adding another convolution layer applied on the last dimension of our log-mel features. Based on our experiments, convolutions with smaller kernels tend to work better, thus we present results with kernel size equal to \num{1}. By adding this convolution layer, we increase the model size only by approximately \SI{2}{\percent} to \num[sepfour]{859983}.

In Table~\ref{EnhancedAcousticFeatures}, we can see that SPICE model-based estimation of pitch (Pitch-SPICE) does not yield an improvement of over YIN estimation (Pitch-YIN). We believe that quality of Pitch-SPICE models can be further improved by retraining the SPICE model on an in-domain audio corpus.

Log-mel features, however, allow us to significantly improve punctuation accuracy. We also observed a slight quality loss for individual punctuation symbols.
We believe this happens because the model is keen to predict more punctuation symbols than it is supposed to based on ground truth labels.

In many of the failure cases, we observed the inherent ambiguity of punctuation prediction.
For example, the sentence \emph{``hey Anna''} can be punctuated in many different ways, as \emph{``hey, Anna!''} or \emph{``hey Anna!''}.
In our current implementation, we do not distinguish such ambiguous cases from clear failures.
In future work, we would like to enhance methods for result interpretation as well as explore further representations of audio.

\end{appendices}
\end{document}